%% file: arxiv.tex
\documentclass[10pt,twocolumn,letterpaper]{article}

\usepackage{iccv}
\usepackage{times}
\usepackage{epsfig}
\usepackage{graphicx}
\usepackage{amsmath}
\usepackage{amssymb}
 \usepackage{subfigure} 

\usepackage[pagebackref=true,breaklinks=true,letterpaper=true,colorlinks,bookmarks=false]{hyperref}
\usepackage{booktabs}
\usepackage{multirow}
\usepackage{multicol}
\usepackage{authblk}

\iccvfinalcopy 

\begin{document}

\title{Adaptive Recursive Circle Framework
	for Fine-grained Action Recognition}

\author[1]{Hanxi Lin}
\author[1]{Xinxiao Wu}
\author[2]{Jiebo Luo}
\affil[1]{Beijing Laboratory of Intelligent Information Technology, School of Computer Science,\protect\\
Beijing Institute of Technology, Beijing 100081, China}
\affil[2]{Department of Computer Science, University of Rochester, Rochester NY 14627, USA  \protect\\
{\tt\small \{hxlin, wuxinxiao\}@bit.edu.cn, jluo@cs.rochester.edu} \vspace{-1ex}
}

\maketitle

\begin{abstract}
	How to model fine-grained spatial-temporal dynamics in videos has been a challenging problem for action recognition. It requires learning deep and rich features with superior distinctiveness for the subtle and abstract motions. Most existing methods generate features of a layer in a pure feedforward manner, where the information moves in one direction from inputs to outputs. And they rely on stacking more layers to obtain more powerful features, bringing extra non-negligible overheads. In this paper, we propose an Adaptive Recursive Circle (ARC) framework, a fine-grained decorator for pure feedforward layers. It inherits the operators and parameters of the original layer but is slightly different in the use of those operators and parameters. Specifically, the input of the layer is treated as an evolving state, and its update is alternated with the feature generation. At each recursive step, the input state is enriched by the previously generated features and the feature generation is made with the newly updated input state.
	  We hope the ARC framework can facilitate fine-grained action recognition by introducing deeply refined features and multi-scale receptive fields at a low cost. Significant improvements over feedforward baselines are observed on several benchmarks. For example, an ARC-equipped TSM-ResNet18 outperforms TSM-ResNet50 with 48\% fewer FLOPs and 52\% model parameters on Something-Something V1 and Diving48.
	
\end{abstract}

\section{Introduction}

\begin{figure*}
	\begin{center}

		\includegraphics[width=1\linewidth]
		{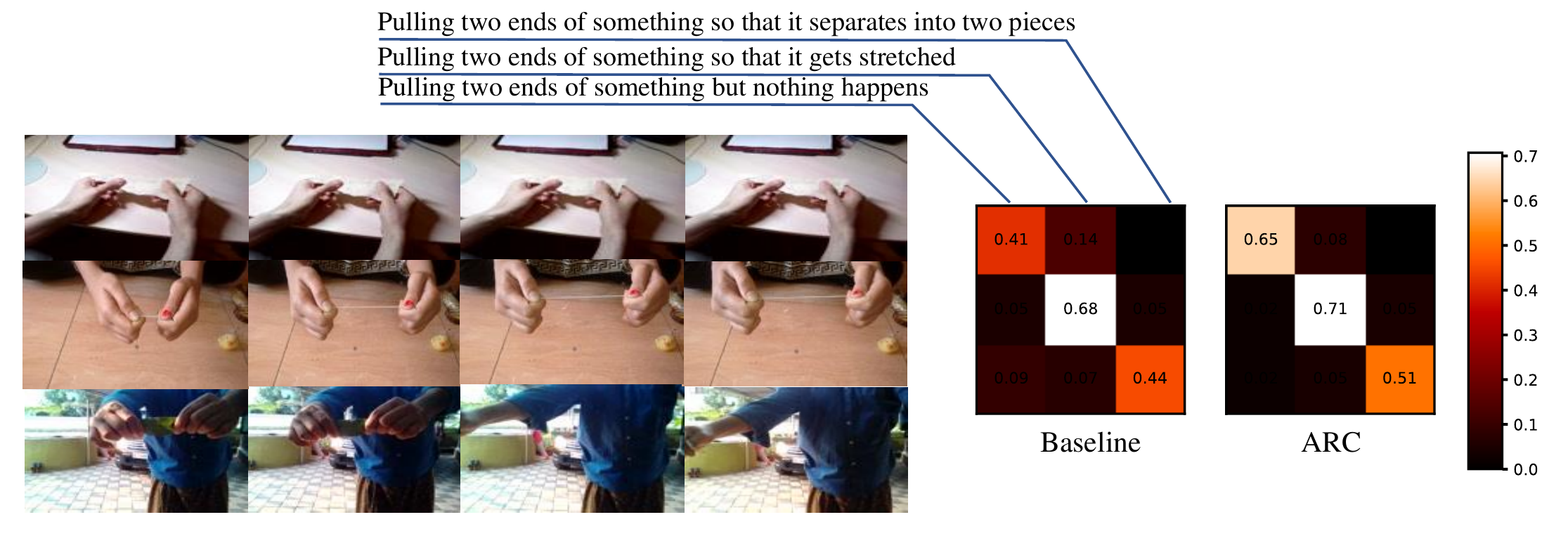}
		\caption{Three fine-grained action categories from the same action group \textit{Pulling two ends of something } in \cite{goyal2017something}. Left: videos; Right: submatrix of the action group taken from the overall confusion matrix. Baseline denotes a TSM-ResNet18 model\cite{lin2017feature} and ARC denotes the ARC equipped TSM-ResNet18 model.}
		\label{fig:Intro}
	\end{center}
\end{figure*}

In action recognition, great progress has been made by introducing end-to-end spatial-temporal CNNs~\cite{tran2015learning,carreira2017quo} and their efficient counterparts~\cite{tran2018closer,xie2018rethinking}. Most of these works are evaluated on benchmarks like 
Sports1M, UCF101
and HMDB51. However, as \cite{li2018resound} shows, these datasets are biased by their scene 
and object appearances, which means that the model could potentially fit the indirect visual cues rather than the actual 
motion. Therefore,  more and more fine-grained datasets~\cite{goyal2017something, li2018resound, shao2020finegym} 
have been proposed as unbiased benchmarks  for action recognition.
To recognize those fine-grained actions, features should capture  spatial-temporal dynamics to distinguish different subtle and abstract motions, rather than representing general scenes or objects. For example, as shown in Fig.\ref{fig:Intro},  recognizing the ``Pulling two ends of something but nothing happens" action requires the  feature rich enough to capture the ``arms or fingers moving oppositely" motion and the ``whether the object is streched  or torn" detail, simultaneously. 

To model the temporal dynamics in the video, some early methods such as 3D CNNs ~\cite{tran2015learning,carreira2017quo} and late temporal fusion models~\cite{wang2016temporal, zhou2018temporal} have been proposed. Many recent methods~\cite{lin2019tsm,luo2019grouped,sudhakaran2020gate,kwon2020motionsqueeze,li2020tea} resort to better temporal modeling on fine-grained spatial-temporal patterns and achieve great success in both accuracy and efficiency.  
Most backbones of these methods are pure feedforward layers (e.g., convolutional layers) where the features of each layer are generated in parallel. Although the feedforward layer is known for its  convenience to the parallel computation, its representation capacity is restricted in the sense that all the basic operations (e.g., convolutions) are implemented on a shared input only once. That is to say, the information moves in one direction from inputs to outputs. Even though deep and rich features can be obtained by stacking more of these feedforward layers, the introduced computation overheads are not desirable. 

In this paper, we propose an Adaptive Recursive Circle (ARC) framework to refine features recursively by forming information movements of adaptive circles between inputs and outputs within a layer, where features become deeper by going through basic operations multiple times  along the path of the adaptive circles. Compared with the normal pure feedforward layers, the ARC augmented layers generate features part by part using  the same basic operation. At each recursive step, the input features are treated as an evolving state and are updated by  backward attention and  adaptive fusion from the previously generated features. The backward attention and the adaptive fusion makes the ARC framework distinct from pure feedforward layers~\cite{szegedy2015going,he2016deep} and Res2Net~\cite{gao2019res2net} style architectures like \cite{li2020tea}. Under the guidance of the generated parts of features, the backward attention aggregates the spatial-temporal information and recursively modulates the input state with different channel attentions to help it focus on different discriminative patterns. Then the adaptive fusion adaptively routes the transformed generated features back to the attended input state to further help it evolve into more powerful representations. This attending and fusion process greatly improves the representation capacity of the original layers. Therefore, the ARC framework brings deeper and richer features with multi-scale receptive fields at a low cost and can be incorporated into all kinds of pure feedforward layers seamlessly.


We evaluate the proposed ARC framework on the Something-Something V1~\cite{goyal2017something}, Diving48~\cite{li2018resound} and Kinetics-400~\cite{kay2017kinetics} datasets. Extensive experiment results show that our ARC outperforms the state-of-the-art methods on fine-grained action recognition. Interestingly, we also observe that an ARC augmented TSM-ResNet-18 backbone achieves better performance than a TSM-ResNet50 backbone with 48\% fewer FLOPs and 52\% fewer parameters on Something-something and Diving48, which verifies the efficient fine-grained modeling of ARC. Our contributions are two-fold:


1. We propose an adaptive recursive circle framework for fine-grained action recognition, which learns deep and rich features by forming information movements of adaptive circles between inputs and outputs within a layer, with  superiority of effectiveness and efficiency. 

2. We propose an adaptive recursive updater as an implement of the recursive information circles using backward attention and adaptive fusion. It works in a sweet spot between efficiency and accuracy and offers great compatibility with most existing CNN backbones and temporal modeling modules.

\section{Related Work}

\subsection{Action recognition}

Deep convolutional neural networks have achieved remarkable success in image recognition.\cite{krizhevsky2012imagenet}.  Later, many researchers investigate how to endow the 2D spatial CNN with powerful temporal modeling capacity for action recognition in videos. Relevant literatures can be summarised into two categories, \emph{i.e.}, motion-encoded handcraft features(such as optical flow) and learnable weighted fusion of features across frames(such as 3D spatial-temporal CNN, 1D temporal convolution and shift operation\cite{lin2019tsm, fan2020rubiksnet,meng2021adafuse}). 

Two-stream Network\cite{simonyan2014two} and DeepVideo\cite{karpathy2014large} are pioneers to incorporate deep data-driven features in action recognition. With 2D CNN, they resort to motion encoded low-level features, \emph{i.e.}, optical flow, and simple fusion strategies across frames respectively for temporal modeling. Tran \emph{et al.} \cite{tran2015learning} first propose the  end-to-end 3D CNN via directly extending 2D spatial kernels with an extra temporal dimension. Carreira \emph{et al.}\cite{carreira2017quo} collect Kinetics dataset and propose to pre-train 3D CNN with inflated pre-trained weights of 2D CNNs to reduce overfitting. Recently, separated spatial convolutions and temporal convolutions\cite{sun2015human,qiu2017learning,xie2018rethinking,tran2018closer,luo2019grouped} are proposed to work more efficient than 3D convolutions. The temporal shift module\cite{lin2019tsm}, a special case of weighted fusion has been proposed to do efficient temporal modeling.  More recently, Fan \emph{et al.}\cite{fan2020rubiksnet} and Meng \emph{et al.}\cite{meng2021adafuse} improve the temporal shift operation by introducing more adaptability to it. Li \emph{et al.} \cite{li2020tea} utilizes temporal differences of feature to model short-term motion and adapts Res2Net block\cite{gao2019res2net} to model long-range temporal dependency.

In this paper, we do not focus on temporal modeling as the above works, but aim to propose a framework of a family of backbones to refine features recursively for action recognition. The proposed ARC layer is designed to be general enough so that it is compatible with most temporal modeling modules.

\subsection{Fine-grained modeling of Action}
Recently, the benchmarks\cite{goyal2017something,li2018resound} of fine-grained video-level action recognition is becoming more and more popular\cite{lin2019tsm,luo2019grouped,sudhakaran2020gate,kwon2020motionsqueeze,li2020tea,li2019temporal,zhu2019approximated}, which feature little bias of indirect static information(e.g., object appearance and background appearance). Some  methods\cite{lin2019tsm,luo2019grouped,sudhakaran2020gate,kwon2020motionsqueeze,li2020tea} resort to better temporal modeling and others\cite{li2019temporal,zhu2019approximated} introduce explicit high-order feature modeling by bilinear operations. While bilinear models obtain high-order features by multiplications between channels, ARC treats the input state of a layer as an evolving state, recursively generates features and recursively updates the input state, which brings higher-order features, implicitly wider network and shows better performance.
In this paper, we propose to tackle the fine-grained modeling of action by introducing adaptive recursion to common pure feedforward feature learning layers. The recursive refinements of the proposed ARC framework within a layer leads to better fine-grained action recognition.

\section{Our Method}
In this section, we present the Adaptive Recursive Circle (ARC) framework to augment pure feedforward layers with the efficient  modeling ability of fine-grained actions. We first start with common feature learning layers that are constructed as pure feedforward networks. Then we introduce the concept and implementations of the ARC framework. Finally, we have discussions on the compatibility of ARC and its relations to existing methods.

\subsection{Pure Feedforward Feature Learning Layers}
Before diving into the proposed ARC, a recap of the common feature layers (e.g., convolutional layer, multi-head self-attention\cite{vaswani2017attention}) is necessary. Most of them are of pure feedforward style. A set of learnable parameters (e.g., convolutional kernels or projection matrix in self-attention) are attached to the layer and responsible for generating each kind of feature from the shared input in parallel. The forward pass of a pure feedforward layer is formulated as
\begin{equation}
\mathbf{Y} = Concat(\{f(\mathbf{X};\mathbf{w}_0),f(\mathbf{X};\mathbf{w}_1),...,f(\mathbf{X};\mathbf{w}_c)\})
\label{conv_formu}
\end{equation}
where $\mathbf{X}$ is the input of the layer, $f(;)$ is a basic operation (e.g., convolution operation) of the layer parameterized by  weight $\mathbf{w}_i$, $c$ is the number of output channels, $Concat()$ denotes  concatenating elements of the set of the input  along the channel dimension, and $\mathbf{Y}$ is the output feature of the layer.

Eq.\ref{conv_formu} shows that the input is refined into the output by applying the basic operations once per layer. Thus, the features of the output are one layer deeper than the input. In the context of fine-grained action recognition, rich and deep features are crucial for drawing conclusions on fine-grained details. Even though more powerful features can be obtained by stacking more such pure feedforward layers, the introduced overheads are not desired in the computationally intensive task of action recognition. It encourages us to propose a framework that efficiently generates the deep and rich features for fine-grained action recognition. 
\begin{figure}
	\begin{center}
		
		
		
		\includegraphics[width=1\linewidth]
		{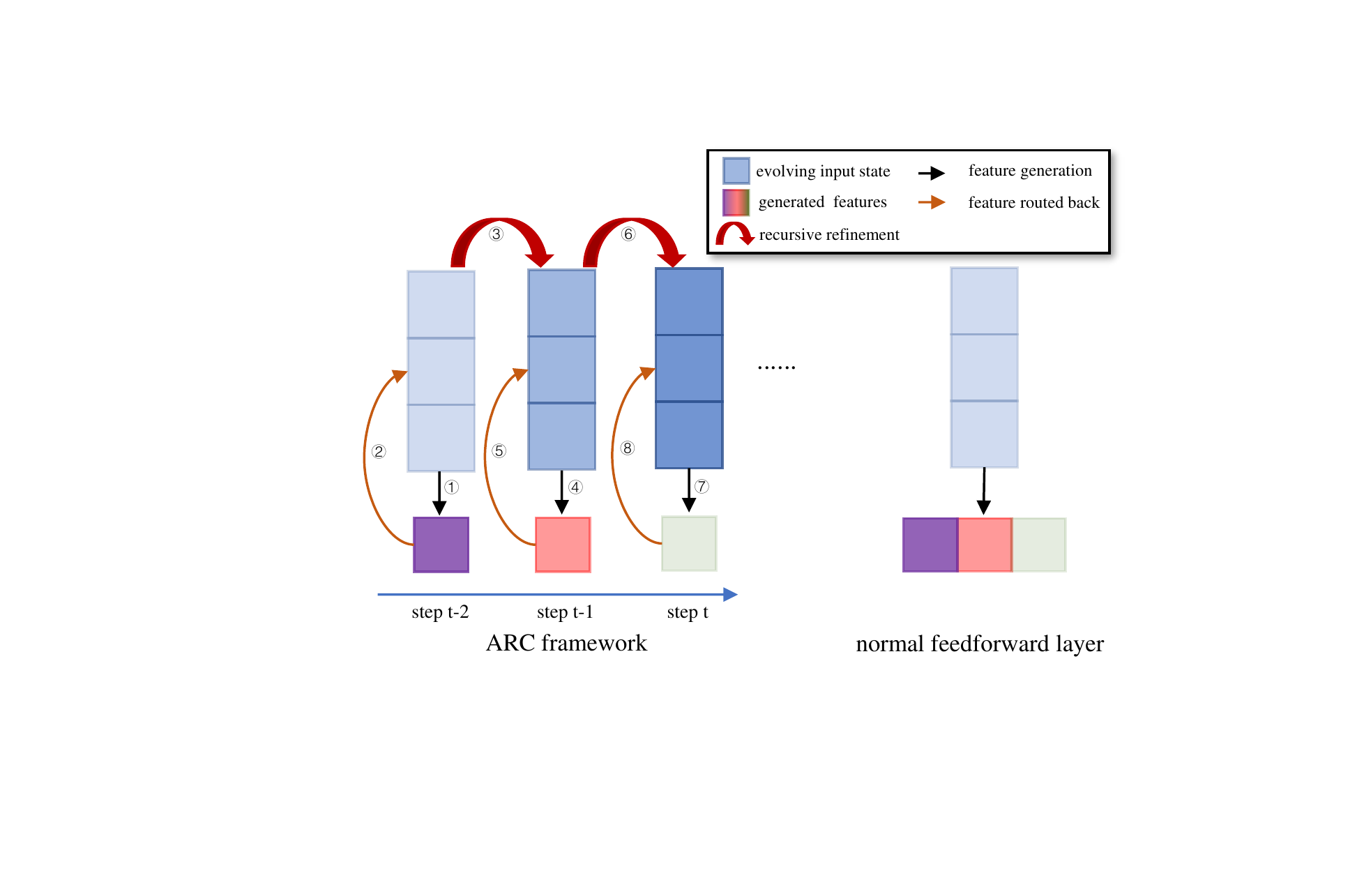}
		\caption{Comparison between ARC framework and normal feedforward layer. ARC adaptively routes generated features back to refine the evolving input state  multiple times. Oppositely, the input of normal feedforward layers is refined only once. For simplicity, some paths of features being routed back in the ARC framework are omitted here. Best viewed in color.}
		\label{ARC}
		
	\end{center}
\end{figure}

\subsection{Adaptive Recursive Circle Framework}
We propose Adaptive Recursive Circle (ARC) framework that learns deep and efficient features by recursively refining features multiple times, as illustrated in Fig.\ref{ARC}. Specifically, different from the pure feedforward style, the feature generation of  the ARC framework is done part by part, which is denoted as \textit{forward path}. And another \textit{backward path} is introduced by treating the input of the layer as an evolving state, and adaptively transmitting the generated features of previous parts to update it. Combining both the forward path and the backward path, the information movement within a layer forms recursive circles between the input and the output,  and the features are refined at each recursive step. The ARC framework is formulated as
\begin{equation}
\mathbf{Y} = Concat(\{\mathbf{Y}_1,\mathbf{Y}_2,...,\mathbf{Y}_n\})
\end{equation}
\begin{equation}
\mathbf{Y}_i = f(\mathbf{Z}_i;\mathbf{w}_i)
\end{equation}
\begin{equation}
\mathbf{Z}_i = \left\{
\begin{array}{ll}
\mathbf{X}_0 &,i=1 \\
ARU(\mathbf{X}_0,\{\mathbf{Y}_{j}\}_{j=1}^{i-1}) &,1<i\leq{n}
\end{array}
\right.
\end{equation}
where $\mathbf{Z}_i$ is the evolving input state, $\mathbf{X}_0$ is the original input of the layer, $n$ is the number of recursive steps, $ARU()$ is the implementation of backward path that will be elaborated in Sec.\ref{ARU}, $f(;)$  serves as the forward path and refers to multiple basic operations here rather than a single one as in Eq.\ref{conv_formu}, and other notations are the same as that in Eq.\ref{conv_formu}.

The ARC framework is a family of feature learning layers, of which the forward path can be instantiated as various feature learning layers and the backward path can also be implemented as different kinds of models. The goal of the ARC framework is to refine the features as much as possible within a feature learning layer to benefit fine-grained action recognition.

\subsection{Adaptive Recursive Updater}
\label{ARU}
There are two core ideas in implementing the adaptive recursive updater (ARU). First, we attend the input state with different attentions at different recursive steps and then adaptively fuse generated features back to the attended input state in the hope of enriching the representation capacity of a layer. Second, we do not want ARU to disturb the initial state of the network and thus the ARC framework can seamlessly leverage the pre-trained weights of original pure feedforward layers. The formulation of ARU is given by
\begin{equation}
\label{ARUeq}
\begin{split}
&ARU(\mathbf{X},\mathbb{F}_{i-1}) = \\ & ReLU(ReLU(\mathbf{X} + \mathbf{W}_e\mathbf{X} + AF(\mathbb{F}_{i-1})+ AM(\mathbb{F}_{i-1})))
\end{split}
\end{equation}
where  $\mathbf{X}$ denotes the original input of the layer,  $\mathbb{F}_{i-1}$ denotes $\{\mathbf{Y}_1,\mathbf{Y}_2,...,\mathbf{Y}_{i-1} \}$, $\mathbf{W}_e\in \mathbb{R}^{C_{in}\times{C_{in}}}$ denotes the embedding matrix that is initialized to all zero, $AF()$ and $AM()$ represent  adaptive fusion  and  backward attention mechanisms, respectively. 
\paragraph{Adaptive Fusion.} A set of weight matrices $\{\mathbf{W}_{fj}\}_{i=1}^n$ are introduced to fuse the output feature into the embedding space, where $\mathbf{W}_{fj}$  $ \in \mathbb{R}^{C_{in}\times{\frac{C_{out}}{n}}}$ is initialized to zero matrix. The adaptive fusion is formulated as
\begin{equation}
AF(\mathbb{F}_{i-1}) = \sum_{j=1}^{i-1}{\mathbf{W}_{fj}\mathbf{Y}_j}
\end{equation}
\paragraph{Backward Attention.} The backward attention mechanism is a kind of channel attention that is based on the global aggregated information. However, different from the common multiplicative attention\cite{hu2018squeeze}, an additive attention operation is utilized for better performance. It works closely with the ReLU activation function in Eq.\ref{ARUeq} as it learns to reduce some channels to the zero-response interval of ReLU function and raise some channels to emphasize them.  The backward attention is formulated as
\begin{equation}
AM(\mathbb{F}_{i-1}) = \sum_{j=1}^{i-1}{\mathbf{W}_{aj}(Pool_S(\mathbf{Y}_j) \oplus Pool_T(\mathbf{Y}_j))}
\end{equation}
where $\mathbf{W}_{aj} \in \mathbb{R}^{C_{in}\times{\frac{C_{out}}{n}}}$, $Pool_S()$ and $Pool_T()$ represent the spatial max pooling and temporal max pooling, respectively, and $\oplus$ denotes summation with broadcasting.

\subsection{Compatibility of ARC Framework}
\paragraph{Compatibility with Backbones and Their Pre-trained Weights.} The only requirement on the compatibility is that the original pure feedforward layers should not use feature grouping since they do not keep full connections between input channels and output channels.  As Sec.\ref{ARU} explains, the ARC framework enhances the feature learning capacity of the original backbone and does not break any connection of the original backbone. Thus the equipment of ARC can be seamlessly performed to enjoy the original pre-trained weights. 
\paragraph{Compatibility with Temporal Modeling Modules.} Since  the ARC framework focuses on fine-grained feature refinement rather than temporal modeling. Most temporal modeling modules\cite{lin2019tsm} and motion modeling modules\cite{kwon2020motionsqueeze} are compatible with the ARC framework, which will be verified in the experiments. 

\subsection{Relations to Existing Models}
\label{reduce}
\paragraph{Relations to Res2net Style Architectures.}The ARC framework brings a new perspective to build deep features with limited overheads by forming adaptive and recursive circles  of information movement between input and output. It is general so that many models can be represented as a reduction of it. For example, when ARU is implemented as parameter-free addition between a group-wisely masked input and output at each recursive step and the incorporated backbones are convolutional layers, the ARC framework turns out to be res2net block\cite{gao2019res2net} and MTA\cite{li2020tea} respectively. 
\paragraph{Relations to Recurrent Models.}The design of the ARC framework is partly inspired by the recurrent hierarchy of LSTM\cite{hochreiter1997long} and other applications of spatial-temporal reasoning\cite{sun2019relational} that is in recurrent style. We incorporate the recurrent style into a feature learning layer of a backbone. The differences lie in the input of the recurrent process and the updating policy. For example, when replacing the update policy of the hidden state and the update policy of the cell state with the basic feature generation operation and ARU, respectively, and feeding the hidden state of last recurrent step to the input of LSTM at the next step,  LSTM becomes an ARC framework.

\section{Experiments}

\subsection{Datasets}
We mainly evaluate the proposed ARC framework on the Something-Something V1~\cite{goyal2017something} and Diving48~\cite{li2018resound} datasets to verify its capacity of fine-grained action recognition. We also conduct experiments on Kinetics400~\cite{kay2017kinetics} to show the generality of the proposed ARC framework. Among these datasets, Something-Something and Diving48 are collected to specifically focus on fine-grained and unbiased~\cite{torralba2011unbiased} action recognition, while action categories in Kinetics400 are more general, coarse-grained, and biased. For example, in Something-Something, ``\textit{pretend to throw something}" and ``\textit{throw something}"  are quite similar in the overall appearance, and only differ in whether the outstretched hand with an object is retracted in the end. In Kinetics400, the action videos, such as ``\textit{playing trumpet}" and ``\textit{bowling}" ,  are collected with general purpose from the internet and the semantics of these actions are often dominated by coarse-grained  and still appearance.  

Something-Something V1 has 174 action categories and 108K videos that vary from 2 to 6 seconds in time, including 86,017 training videos, 11,522 validation videos, and 10,960 test videos. We report the top-1 and top-5 accuracy on the validation set and test set.

Diving48 includes 18,404 trimmed video clips of 48 unambiguous dive classes. Videos are split into training and validation set. We report top-1 accuracy on the validation set. 

Kinetics400 has $\sim$240K training videos and $\sim$20K validation videos, covering 400 action categories. These videos are collected from Youtube. Since some video URLs are no longer available,  the actual number of videos used  varies in the literature. In our experiments, we have 240,436 training videos and 19,404 validation videos. We report top-1 and top-5 accuracy on the validation set.
\subsection{Implementation Details}
\paragraph{Backbone.}  We choose ResNet~\cite{he2016deep} equipped with temporal modeling modules~\cite{lin2019tsm,kwon2020motionsqueeze} as our backbone and evaluate the effectiveness of the proposed method by augmenting the convolutional layers with ARC, except for those 1$\times$1 point-wise convolutional layers. 

\paragraph{Data Augmentation.} For Something-Something V1 and Diving48, frames are first scaled to size 240$\times$320, following~\cite{zolfaghari2018eco,kwon2020motionsqueeze}. For Kinetics400, we scale the shorter size of the video to 320 pixels. Then random scaling and cropping of size 224$\times$224  are adopted throughout these three datasets. The random horizontal flip is also used. Note that in Something-Something V1,  random horizontal flip is applied selectively, following~\cite{sudhakaran2020gate}. 

\paragraph{Training and Testing.} We  adopt sparse sampling strategy~\cite{wang2016temporal} across frames. All models are pre-trained from ImageNet pre-trained weights. We use SGD as the optimizer and apply dropout before the classification head. For Something-Something V1 and Kinetics400, we apply  multi-step learning rate schedule. The learning rate decay happens at $30_{\text{th}}$, $40_{\text{th}}$ epoch by a factor of 10 for both of them, and one more decay happens at $45_{\text{th}}$ for Kinetics400. For Diving48, we adopt a cosine learning rate schedule, following ~\cite{sudhakaran2020gate}. The first 10 epochs are for gradual warm-up~\cite{goyal2017accurate}. For all the datasets, the number of recursive steps of ARC is set to 4, unless otherwise specified. Other training hyper-parameters are summarized in Table~\ref{tab:hyper}.
\begin{table}[htbp]
	\begin{center}
		\caption{\label{tab:hyper} Training hyper-parameters.}
		\scalebox{0.60}{
			\begin{tabular}{cccccccc}   
				\toprule    
				dataset   & weight decay & batch size & epochs & initial lr & dropout rate \\  
				\midrule
				Something-Something V1  & 5e-4 & 32  &50 & \multirow{3}{*}{0.01} & \multirow{3}{*}{0.5}\\
				
				Diving48 & 5e-4 & 8 &30 & ~  & ~\\
				Kinetics400  & 1e-4 & 64 &50 & ~  & ~\\
				\toprule 
				
			\end{tabular} 
		}
	\end{center}
\end{table}

\subsection{Experimental Results}

\subsubsection{Performance of ARC on Improving Feedforward Models}
We apply the ARC framework to TSM~\cite{lin2019tsm} backboned by ResNet, and conduct a fair comparison between the baselines with or without ARC augmentation in the same training setting on Something-Something V1 and Diving48. As a brief introduction for the baseline, TSM employs a simple temporal shift operation to the spatial-temporal feature map to enjoy the 3D convolution while keeping the computational budget constant. It relies on pure feedforward feature extractors (2D CNNs) as its backbone, including ResNet18 and ResNet50.

The results are reported in Table~\ref{tab:improve}. We have several observations as follows. First, our method achieves significant improvements  over TSM in all cases, which validates the superiority of the proposed ARC framework on fine-grained action recognition. Second, the performance boost is achieved by introducing minor FLOPs and parameters. The ARC augmented models backboned by ResNet18 approximate or even surpass their normal ResNet50 counterpart, with nearly half FLOPs and parameters. This encouraging results highlight the superiority of deeper features at various scales on fine-grained action recognition which is modeled by an adaptive backward path. Third, significant improvements (more than 3\%) are also observed in a larger backbone (e.g., ResNet50), suggesting that our ARC framework scales well from small backbones to large backbones on fine-grained action recognition.

\begin{table}[htbp]
	\label{improve}
	\begin{center}
		\caption{\label{tab:improve}Augmenting ARC framework with different models and backbones. Performance is measured by top-1 accurary on fine-grained action recognition. }
		\scalebox{0.65}{
			\begin{tabular}{cccccccc}   
				\toprule    
				~ & Model & Backbone & FLOPs(G) & Param.(M) & Top-1(\%) &Gain\\  
				\toprule 
				
				\multirow{4}{*}{\rotatebox{90}{Something}} & TSM(reproduced) & ResNet18 &14.6  & 11.27  & 41.8 & \multirow{2}{*}{+6.1}\\ 
				~ & ARC-TSM & ResNet18 &17.2  & 14.2  & \textbf{47.9} & ~\\  
				\cmidrule{2-7}
				~ & TSM(reproduced) & ResNet50 &33  & 24.3  & 47.8 & \multirow{2}{*}{+3.4}\\ 
				~ & ARC-TSM & ResNet50 &37.2  & 27 & \textbf{51.2} & ~\\

				\midrule \midrule 
				\multirow{4}{*}{\rotatebox{90}{Diving48}} & TSM(reproduced) & ResNet18 &58.4  & 11.27  & 36.0 & \multirow{2}{*}{+3.6}\\ 
				~ & ARC-TSM & ResNet18 &68.8  & 14.2  & \textbf{39.6} & ~\\  
				\cmidrule{2-7}
				~ & TSM(reproduced) & ResNet50 &132  & 24.3  & 38.8 & \multirow{2}{*}{+3.6}\\ 
				~ & ARC-TSM & ResNet50 &148.8  & 27 & \textbf{42.4} & ~\\
				\toprule
			\end{tabular} 
		}
	\end{center}
\end{table}

\begin{table*}[htbp]
	\begin{center}
		\caption{\label{tab:sthV1}Performance comparison on Something-Something V1. Results marked by $\dagger$ are from our implementations.}   
		\scalebox{0.7}{
			\begin{tabular}{ccccccc}   
				\toprule   \textbf{Model} & \textbf{Backbone} & \textbf{Pre-train} & \textbf{Frames} & \textbf{FLOPs$\times{}$clips/G} &\textbf{Param./M} & \textbf{Val Top-1/Top-5(\%)} \\
				\midrule
				2D CNN + late fusion: & ~ & ~ & ~ & ~ & ~ & ~ \\
				TSN\cite{wang2018temporal} from \cite{lin2019tsm} & ResNet50 & ImgNet & 8 & 33G$\times{}$1 & 24.3 & 19.7/- \\  
				TRN\cite{zhou2018temporal} & BN-Inception & ImgNet & 8 & 8$\times{}$N/A  & 10.5 & 34.4/- \\  
				\midrule
				Temporal modeling modules: & ~ & ~ & ~ & ~ & ~ & ~ \\
				TSM-RGB\cite{lin2019tsm} & ResNet50 & K400 & 16 & 65$\times{}$1 & 24.3 & 47.3/77.1 \\ 
				TSM-RGB\cite{lin2019tsm}$\dagger$ & ResNet101 & ImgNet & 8 & 63.1$\times{}$1 & 42.9 & 50.0/78.7 \\ 
				STM\cite{jiang2019stm}& ResNet50 & ImgNet & 16 & 67$\times{30}$ & 24.0 & 50.9/80.4 \\
				GSM\cite{sudhakaran2020gate} & InceptionV3 & ImgNet & 16 & 54$\times{}$2  & 22.2 & 51.7/-\\  
				TPN\cite{yang2020temporal} & TSM-ResNet50 & ImgNet & 8 & N/A$\times{10}$ & - & 49.0/- \\
				CorrNet-R101\cite{wang2020video} & ResNet101 & - & 32 & 187$\times{10}$ & - & 50.9/-\\
				MSNet\cite{kwon2020motionsqueeze} & TSM-ResNet50 & ImgNet & 16 & 67$\times{}$1 & 24.6 & 52.1/82.3 \\
				\midrule
				Spatial-temporal backbones: & ~ & ~ & ~ & ~ & ~ & ~ \\
				ECO \cite{zolfaghari2018eco} & BN-Inc+3D ResNet18 & K400 & 32 & 92$\times{}$1 & - & 46.4/-  \\  
				I3D from \cite{wang2018non} & 3D ResNet50 & \multirow{2}{*}{ImgNet+K400} & 32 & 153$\times{}$2 & 28.0 & 41.6/72.2 \\  
				NL-I3D from \cite{wang2018non} & 3D ResNet50 & ~ & 32 & 168$\times{}$2 & 35.3 & 44.4/76.0 \\
				S3D-G \cite{xie2018rethinking} & BN-Inception+I3D & ImgNet & 64 & 71$\times{}$1 & 11.6 & 48.2/78.7 \\  
				ABM-iabp\cite{zhu2019approximated}$\dagger$ & ResNet18 & ImgNet & 8$\times{}$3 & 30.4$\times{}$1 & 26.8 & 45.1/74.0\\
				$\text{TEA}$\cite{li2020tea} & ResNet50 & ImgNet & 16 & 70$\times{}$30 & N/A & 52.3/81.9 \\  
				$\text{RubiksNet}$\cite{fan2020rubiksnet} & ResNet50 & ImgNet & 8 & 16$\times{}$1 & 8.5 & 46.4/74.5 \\
				\midrule
				$\text{ARC}$(ours) & TSM-ResNet18 & ImgNet & 8 & 17$\times{}$1 & 14.2 & 47.9/76.4\\
				$\text{ARC}$(ours) & $\text{TSM}$-ResNet50 & ImgNet & 8 & 37$\times{}$1 & 27.0 & 51.2/79.0\\
				$\text{ARC}$(ours) & $\text{TSM}$-ResNet50 & ImgNet & 16 & 74$\times{}$1 & 27.0 & 53.4/81.8\\
				$\text{ARC}$(ours) & $\text{TSM}$-ResNet50 & ImgNet & 8+16 & 111$\times{}$1 & 54.0 & \textbf{55.0}/82.6\\
				
				\toprule
			\end{tabular} 
		}
	\end{center}
\end{table*}

\subsubsection{Comparisons with State-of-the-art Method}

Table~\ref{tab:sthV1} summarizes the results on Something-Something V1. We divide the existing methods into three categories: 2D CNNs with late fusion, temporal modeling module, and spatial-temporal backbones. The methods of the first category(the first part of the table) only incorporate temporal modeling on top of features of 2D CNNs after spatial pooling and fine-grained modeling of spatial-temporal information is beyond their capacity, resulting in low performance. The methods of the other two categories( the second and third parts of the table incorporate temporal modeling across different stages of networks, thus obtaining  higher performance. However, these methods(except for TEA~\cite{li2020tea}), adopt pure feedforward feature extractors(i.e., normal convolutional layers). Among them, MSN\cite{kwon2020motionsqueeze} backboned by TSM-ResNet50 of 16 frames achieve the best performance. Our ARC model with the same backbone achieves  1.3\% higher accuracy. Note that the only thing that what ARC framework does is adding recursive backward paths to TSM, without any further temporal modeling. These observations demonstrate the simplicity and the effectiveness of the ARC framework on fine-grained action recognition. Since the FLOPs of the compared methods vary a lot, we also report the ARC models backboned by TSM-ResNet18 and MSN-TSM-ResNet18. Compared with the most efficient method, i.e. RubiksNet\cite{fan2020rubiksnet}, our ARC models achieve 1.5\% and 3.1\% higher top-1 accuracy with 1G and 2G more FLOPs, respectively. 

Table~\ref{tab:diving} summarizes the results on Diving48 V1. Similar to the observations on Something Something. The ARC framework augmented TSM-ResNet18 model outperforms TSM-ResNet50 by 0.8\% by nearly half FLOPs and parameters. It also surpasses other SoTA methods with much fewer overheads, except GSM. Furthermore, the TSM-ResNet50-based ARC model achieves SoTA results with 42.4\% top-1 accuracy on Diving48. For Diving48 V2, results are shown in Table~\ref{tab:divingv2} of the appendix.
\begin{table*}[htbp]
	\begin{center}
		\caption{\label{tab:diving}Performance comparison on Diving48 V1}
		\scalebox{0.7}{
			\begin{tabular}{ccccccc}   
				\toprule   \textbf{Model} & \textbf{Backbone} & \textbf{Pre-train} & \textbf{Frames} &$\textbf{FLOPs/G}\times$\textbf{ clips} &\textbf{Param./M} & \textbf{Val Top-1(\%)} \\  
				\midrule  \midrule  
				Att-LSTM\cite{kanojia2019attentive} & ResNet18 & ImgNet & 64 & N/A$\times{1}$ & - & 35.6\\
				$\text{GST}$\cite{luo2019grouped} & ResNet50 & ImgNet & 16 & 59$\times{1}$ & 21 & 38.8 \\
				$\text{TSM}$\cite{lin2019tsm}(our impl.) & ResNet18 & ImgNet & 16 & 29.15$\times{2}$ & 11.27 & 36.0 \\
				$\text{TSM}$\cite{lin2019tsm}(our impl.) & ResNet50 & ImgNet & 16 & 65$\times{2}$ & 24.3 & 38.8 \\
				CorrNet-R101\cite{wang2020video} & ResNet101 & - & 32 & 187$\times{10}$ & - & 38.2\\
				$\text{GSM}$\cite{sudhakaran2020gate} & IncV3 & ImgNet & 16 & 54$\times{2}$ & - & 40.3 \\
				\midrule
				\midrule  $\text{ARC}$(ours) & TSM-ResNet18 & ImgNet & 16 & 34.2$\times{2}$ & 14.2  & 39.6 \\
				$\text{ARC}$(ours) & TSM-ResNet50 & ImgNet & 16 & 74.2$\times{2}$ & 27.0 & \textbf{42.4} \\
				\toprule
			\end{tabular} 
		}
	\end{center}
\end{table*}

Table~\ref{tab:kinetics} summarizes the results on Kinetics400. Even though Kinetics400 is not a fine-grained dataset. Our ARC framework still achieves comparable results with SoTA methods, showing the generality of ARC. 

\subsubsection{Ablation Analysis}
We conduct extensive experiments to investigate the design of ARU in detail and vary several hyper-parameters to show some features of our ARC framework. All these experiments of ablation analysis are conducted on Something-Something V1 and top-1 validation accuracy is reported. Unless otherwise specified, the backbone, the number of recursive steps and the number of sampled frames are fixed to TSM-Resnet18, 4 and 8, respectively.
\paragraph{Ablations on ARU.} We remove different components of ARU and evaluate the resultant performance. Results are summarized in Table~\ref{tab:ARU}. Interestingly, we found that while the feature resolution of the backward attention is  damaged greatly by the pooling operations, using backward attention alone still works better than using adaptive fusion alone. It suggests that the seperated design of spatial-temporal pooling preserves enough spatial-temporal structures while reducing computational burdens. Overall, all components contribute to the improvements, especially for the backward attention. 
\paragraph{Number of Recursive Steps.} We tune the number of recursive steps $n$ to see the performance trends with respect to $n$. The results are shown in Table~\ref{tab:steps}. We observe that the performance improves when $n$ is larger. It is intuitive that larger $n$ brings longer recursions in ARC and generats deeper features, thus benefiting fine-grained action recognition. It is worth noting that the overheads in terms of FLOPs and the number of parameters are constant after $n$  n exceeds 2. But, as shown in the overhead analysis in the appendix, the memory consumption grows linearly with the increasing $n$.
\paragraph{Depths of ARC Augmentation.} We apply the ARC framework to different stages of ResNet and the results are illustrated in Table~\ref{tab:depths}. We first apply the ARC framework to every single stage and find that the ARC framework brings consistent improvements and works better in deeper stages. It is probably due to that features at deeper layers include richer semantics and ARC is capable of adaptively leveraging this rich infomation.  It is worth noting that even though we show that the ARC framework brings improvements across all stages of the backbone, the recursive hierarchy of ARC is memory-consuming. Thus we only use the ARC framework in the last three stages of relatively low dimensions in the rest of the paper. 
\paragraph{Backward Attention in ARU.} We investigate the three aspects(i.e., way of interaction, information aggregation and layer that generates attention weights) of the backward attention in ARU. The results are reported in Table~\ref{tab:Atten}. For the way of interaction, we compare multiplication and simple addition and the results show that additive attention outperforms multiplicative attention, probably due to the diminished gradient problem of sigmoid function before the multiplication. It is more difficult to learn the drastic suppression and activation compared with a simply addition, thus preventing ARC from generating more diverse features during the recursion. For the information aggregation, we compare several strategies: spatial pooling on each frame(denoted as ``S''), spatial-temporal pooling across all spatial-temporal positions (denoted as ``ST''), temporal pooling across all spatial positions (denoted as ``T''), and broadcasting summation of separated spatial pooling and temporal pooling (denoted as ``S+T''). It is obvious that ``S+T'' performs best since it preserves much more coordinate information. For the layer that generats attention weights, we compare GRU with a simple fully connected layer(denoted as ``FC''). GRU takes each part of generated features as input and outputs its hidden states as the attention. The results show that the fully connected layer performs much better, probably due to that the implementation of GRU is over-parameterized.

\begin{table}[htbp]
	\label{ablation}
	\caption{Ablation study.}
	\begin{center}
	\subtable[Ablations on ARU.]{
	\label{tab:ARU}
	\scalebox{0.67}{
		\begin{tabular}{ccc|c}   
			\toprule    
			linear embedding & backward attention & adaptive fusion &  Top-1(\%)\\  
			\toprule 
			~ & ~ & \checkmark   & 45.3 \\ 
			~ & \checkmark & ~   & 45.9 \\ 

			\checkmark &~ & \checkmark   & 46.3\\ 
			\checkmark &\checkmark & ~  & 46.7\\ 
			~ & \checkmark & \checkmark   & 46.9\\
			 \checkmark & \checkmark & \checkmark   & ~\textbf{47.9}\\ 
			\toprule
		\end{tabular} 
	}
	}
	\subtable[Performance comparison on different number of recursive steps.]{
	\label{tab:steps}
	\scalebox{0.75}{
		\begin{tabular}{ccc|c}   
			\toprule    
			$n$ & FLOPs/G & Param./M  & Top-1(\%)\\  
			\toprule 
			1 &  14.6 & 11.27  & 41.8\\ 
			2 & 17.2  & 14.2  & 46.5\\  
			4 & 17.2 & 14.2 & \textbf{47.9}\\ 
			\toprule
		\end{tabular} 
	}
	}
		\subtable[Ablations on different depths of ARC augmentation.]{
		\label{tab:depths}
		\scalebox{0.75}{
			\begin{tabular}{ccc|c}   
				\toprule    
				Augmented Stages & FLOPs/G & Param./M  & Top-1(\%)\\  
				\toprule 
				- & 14.57  & 11.27  & 41.8\\ 
				res2 & 15.37  & 11.31  & 43.8\\ 
				res3 & 15.44  & 11.41  & 44.9\\ 
				res4 & 15.43  & 11.82  & 45.9\\ 
				res5 & 15.43  & 13.49  & 45.9\\
				res4,5 & 16.29  & 14.05  & 47.1\\  
				res3,4,5 & 17.16 & 14.19 & \textbf{47.9}\\ 
				\toprule
			\end{tabular} 
		}
	}
		\subtable[Ablations on designs of backward attention.]{
		\label{tab:Atten}
		\scalebox{0.75}{
			\begin{tabular}{ccc|c}   
				\toprule    
				Interaction & Aggregation & Layer & Top-1(\%)\\  
				\toprule 
				
				\multirow{2}{*}{Multiplicative} & S  & GRU    & 45.0 \\ 
				~  & S & FC  & 46.2 \\  
				\midrule
				\multirow{5}{*}{Additive}  & S & GRU   & 45.6 \\  
				~  & ST & FC & 46.5 \\  
				~ & T & FC  & 47.5 \\ 
				~ & S & FC   & 47.0 \\ 
				~  & S+T & FC  & \textbf{47.9} \\
				\toprule
			\end{tabular} 
		}
		}

	\end{center}
\end{table}


%
%

\begin{table}[htbp]
	\begin{center}
		\caption{\label{tab:kinetics}Performance comparison on Kinetics400 validation set.}
		\scalebox{0.65}{
			\begin{tabular}{ccccc}   
				\toprule   \textbf{Model} & \textbf{Backbone} &Frames &$\textbf{FLOPs/G}\times$\textbf{clips}  & \textbf{Val Top-1(\%)} \\  
				\midrule \midrule 
				TSN\cite{wang2016temporal} & InceptionV3  &25 & 3.2$\times{250}$  & 72.5 \\
				\midrule \midrule 
				MSN\cite{kwon2020motionsqueeze} & TSM-ResNet50 &8  & 34$\times{10}$  & 75.0 \\
				TSM\cite{lin2019tsm} & ResNet50  &8 & 33$\times{30}$  & 74.1 \\
				\midrule \midrule 
				I3D\cite{carreira2017quo} & BN-Inception & 64  & 108$\times{}$N/A  & 72.1 \\
				R(2+1)D\cite{tran2018closer} & ResNet34 & 32 &152$\times{}$10 &74.3\\
				SlowOnly\cite{feichtenhofer2019slowfast} & ResNet50 &8  & 41.9$\times{10}$  & 74.8 \\
				TEA\cite{li2020tea} & ResNet50 &8  & 35$\times{10}$  & 75.0 \\
				
				\midrule
				\midrule 
				$\text{ARC}$(ours) & TSM-ResNet50  &8 & 37$\times{30}$  & 75.0 \\
				\toprule
			\end{tabular} 
		}
	\end{center}
\end{table}

\begin{figure*}
	\begin{center}
		
		\includegraphics[width=1\linewidth]
		{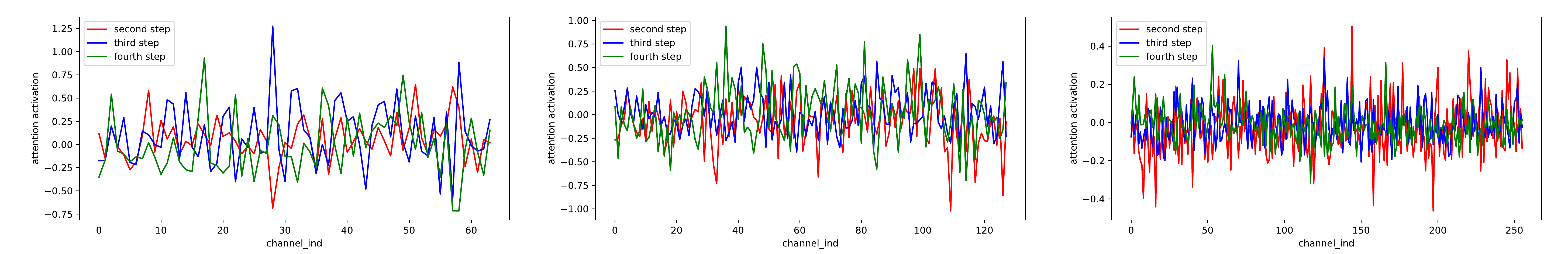}
		\caption{Visualization of backward attention activations at different recursive steps and different positions. Left: stage2-block0-layer0, Middle: stage3-block0-layer0, Right: stage4-block0-layer0.}
		\label{vis:atten_steps}
	\end{center}
\end{figure*}

\subsection{Interpretation}
We offer the visualization of backward attentions here. More interpretations are available in the appendix.
\paragraph{Attentions at Different Recursive Steps.} The process of recursive refinement is a key feature of the proposed ARC framework. Thus  we would like to see how ARU works at different recursive steps in detail. We visualize attention activation since it is more convenient to visualize a vector shared among all spatial-temporal positions. The results are shown in Fig.\ref{vis:atten_steps}. It is interesting to observe that the attention from the same layer vary a lot in different recursive steps, which demonstrates that the backward attention probably learns to suppress and activate each channel in different ways at different steps, thus helping the input state evolve into more powerful and more diverse representations. This argument is also supported by the visulization of the evolving input state in the appendix.
%
%


\section{Conclusion}
We have presented an adaptive recursive circle framework for fine-grained action recognition, which refine features recursively by forming adaptive circles of forward and backward paths between inputs and outputs within a layer. Our method is able to efficiently learn deeper and richer features with multi-scale receptive fields, thus succeeding in distinguishing different subtle and abstract motions for recognition. Our method is a general  framework that can be readily and easily applied to most existing pure feedforward backbones. Extensive experiments on several benchmarks demonstrate the effectiveness of the proposed  adaptive recursive circle framework  on fine-grained action recognition. In the future, we are going to extend our method to more network architectures such as Transformer. 

{\small
\bibliographystyle{ieee_fullname}
\bibliography{egbib}
}

\input{text/supple_in_mainbody}

\end{document}

%% file: text/supple_in_mainbody.tex
\newpage
\appendix
\newpage







\section{Interpretation}
In this section, we investigate the interpretation of ARC in several aspects, including the confusion matrix, the class activation mapping and the input state evolving. TSM-ResNet18 is chosen as the baseline and the ARC-equipped TSM-ResNet18 is chosen to be an instantiation of our proposed ARC framework. The number of recursive steps $n$ is set to 4.

\subsection{Confusion Matrix.}
We compare different submatrices in the overall confusion matrix to evaluate the ability of fine-grained action recognition on Something-Something V1. Concretely, based on the action groups defined in \cite{goyal2017something}, we take out several rows and columns from the confusion matrix that correspond to each category within a certain action group. Each element in the submatrix represents the probability of classifying the category of the row into the category of the column. 

\begin{figure*}
	\begin{center}
		\includegraphics[width=0.9\linewidth]
		{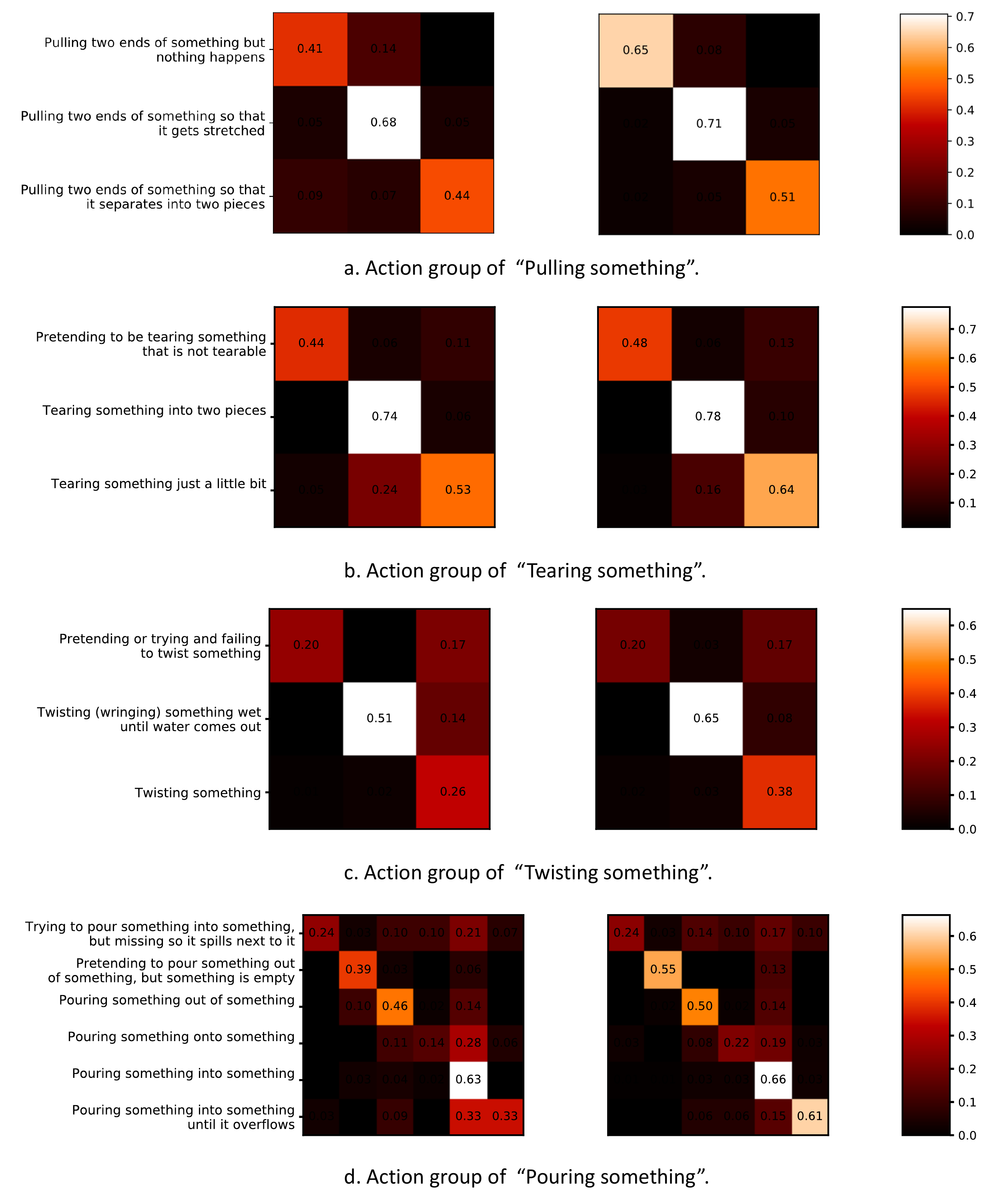}
		\caption{Confusion matrices of several action groups in \cite{goyal2017something}. Left: baseline, Right: ARC-equipped baseline.}
		\label{fig:confusion}
		
	\end{center}
\end{figure*}

We show examples of several action groups in Something-Something V1 in Fig.\ref{fig:confusion}. Each action group contains fine-grained action categories that are similar to each other. Wrong predictions (not on the main diagonal) are greatly reduced and correct predictions (on the main diagonal) are increased in ARC-equipped baseline, suggesting that the ability of fine-grained modeling is greatly improved by equipping the proposed ARC framework.

\subsection{Class Activation Mapping.}
We visualize Class Activation Mapping (CAM) of baseline and the ARC framework, using Grad-CAM\cite{selvaraju2017grad} based on the features of Res3 layer. We visualize videos of five different categories. 

\begin{figure*}
	\begin{center}
		\includegraphics[width=1\linewidth]
		{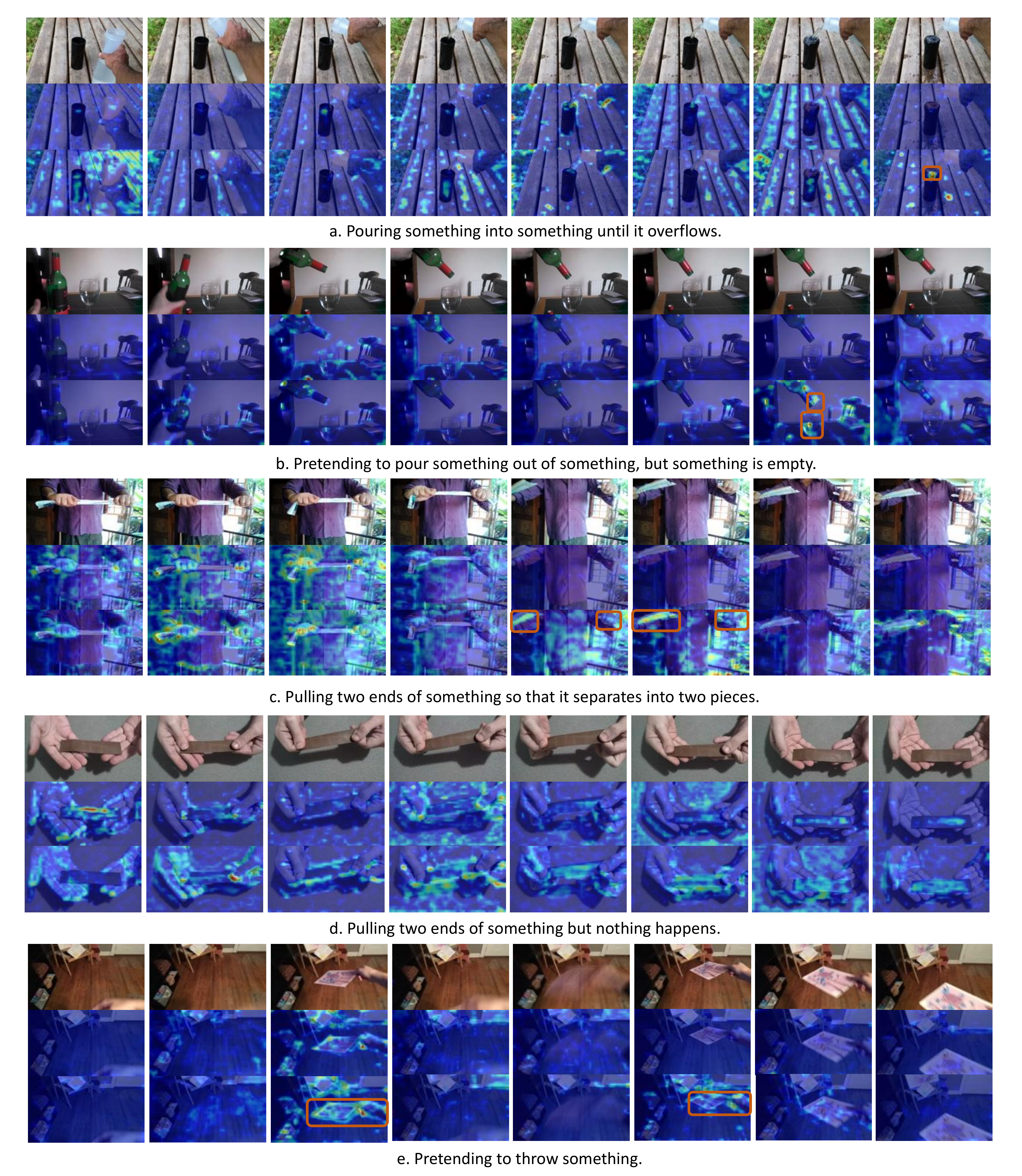}
		\caption{Class activation mapping of actions of five different action categories in \cite{goyal2017something}. For each action, the first row represents raw RGB video, and the second and third row represent the CAM of baseline and the ARC-equipped baseline, respectively. Orange boxes denote that fine-grained and discriminative patterns are captured by the ARC framework.}
		\label{fig:cam}
	\end{center}
\end{figure*}

The results are shown in Fig.\ref{fig:cam}. For the action of ``Pouring something into something until it overflows", the ARC framework captures the fine-grained detail that the water overflows, thus not misclassifying the action into ``Pour something into something". Similar observations are found in Fig.\ref{fig:cam}.b, Fig.\ref{fig:cam}.c and Fig.\ref{fig:cam}.e too, where the ARC framework recognizes ``nothing poured into the goblet", ``two pieces of something" and ``throwing is withdrawn", respectively. In Fig.\ref{fig:cam}.d, the action ``Pulling two ends of something but nothing happens" features ``nothing happens". The ARC framework shows strong interests in the boundaries of hands and object. We conjecture that the ARC framework is trying to distinguish whether something is happened to the hands and objects by focusing on the fine-grained movements around the boundaries of them.

\subsection{Input State Evolving.}
In the ARC framework, the input state of a layer is evolving during the recursion. Here, we visualize the evolving process to show how this design works, based on the second 3$\times{}$3 convolutional layer of the second residual block of Res3 features. To visualize the feature maps as normal 2D images, we average them across channels.

\begin{figure*}
	\begin{center}
		\includegraphics[width=1\linewidth]
		{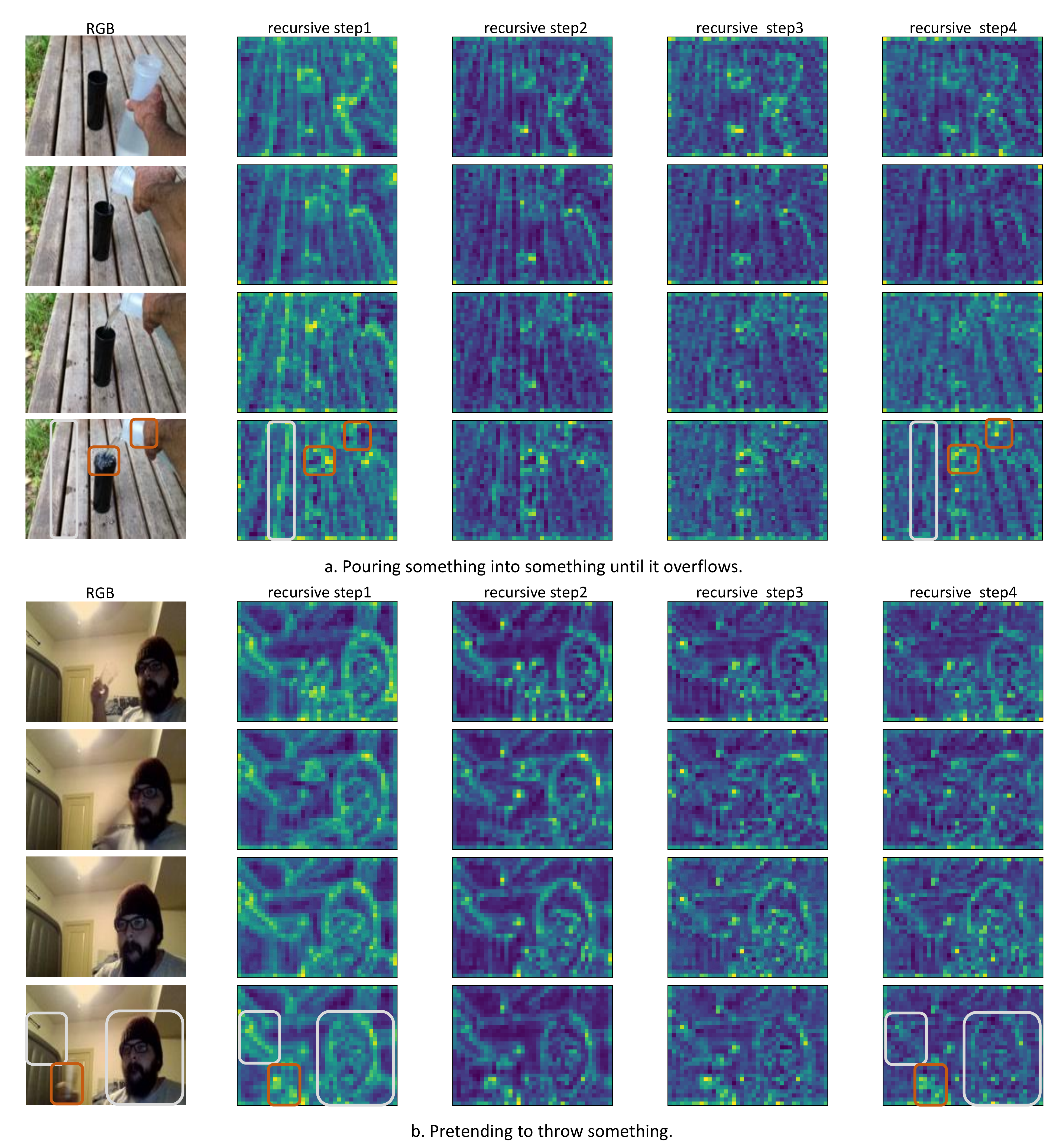}
		\caption{Visualization of the evolving process of the input state in the ARC framework. The input states of all $n$ recursive steps are averaged across channels for visualization, reflecting the degree of excitation of features in different regions. The ARC framework partly suppresses the irrelevant appearances of object and scene (represented by grey boxes), and enhances the fine-grained action-related details during the recursion (represented by orange boxes).}
		\label{fig:evolving}
	\end{center}
\end{figure*}

As shown in Fig.\ref{fig:evolving}, the feature maps at the first recursive step, or rather, the original input of the layer, shows clear interests in every object of both the foreground and the noisy background (e.g., the gap between the planks). But when the recursion goes on, the irrelevant object features are greatly suppressed. At the end  of the recursion, the ARC framework mainly focuses on fine-grained and discriminative patterns (e.g., the overflowing water and the withdrawn hand with object). 

It is also worth noting that the recursion of ARC does not always work. For example, in the first sample frame of Fig.\ref{fig:evolving}.b, the ARC framework ends up looking at irrelevant objects of the background. It might be due to the overfitting of extracting noises of background objects.

%

\section{Compatibility with Temporal Modeling Modules}
As we mentioned in the main body, rather than explicitly modeling temporal dynamics, our ARC refines the given features recursively to facillitate  modeling of any fine-grained pattern, which is agnostic to temporal modeling. Thus, improvements are expected when further applying the ARC framework to the models of temporal modeling. In this section, we equip backbones with both the ARC framework and temporal modeling modules to investigate the compatibility of ARC with temporal modeling modules. Concretely, TSM\cite{lin2019tsm} and MS Module (MSM)\cite{kwon2020motionsqueeze} are used as representatives for temporal modeling modules. The number of recursive steps $n$ is set to 4. The number of sampled frames is set to 8. Performance is measured by top-1 accuracy on the validation set of Something-Something V1. Training and testing are done under the same setting. 

The results are shown in Table~\ref{tab:supp_improve}. ARC significantly improves TSM-ResNet18 and TSM-ResNet50 by 6.1\% and 3.4\%, respectively, showing that ARC is largely orthogonal to TSM. When the temporal modeling is further enhanced by MSM, the improvements over ResNet18 and ResNet50 are 3.3\% and 1.5\%, respectively. The improvements are lower but still evident, suggesting that the model capacity of ARC and MSM overlap slightly. In summary, ARC is well compatible with temporal modeling modules. In other words, further improvements are expected when applying ARC to more SoTA models that are designed for temporal modeling.

\newpage
\clearpage

\newpage
\clearpage

\newpage
\clearpage

\newpage
\begin{table}[ht!]
	\begin{center}
		\caption{\label{tab:supp_improve}Compatibility with Temporal Modeling Modules. }
		\scalebox{0.6}{
			\begin{tabular}{cccccccc}   
				\toprule    
				~ & Model & Backbone & FLOPs (G) & Param. (M) & Top-1 (\%) &Gain\\  
				\toprule 
				
				\multirow{4}{*}{\rotatebox{90}{TSM\cite{lin2019tsm}}} & TSM (reproduced) & ResNet18 &15  & 11.3  & 41.8 & \multirow{2}{*}{+6.1}\\ 
				~ & ARC-TSM & ResNet18 &17  & 14.2  & \textbf{47.9} & ~\\  
				\cmidrule{2-7}
				~ & TSM (reproduced) & ResNet50 &33  & 24.3  & 47.8 & \multirow{2}{*}{+3.4}\\ 
				~ & ARC-TSM & ResNet50 &37  & 27 & \textbf{51.2} & ~\\
				
				\midrule \midrule 
				\multirow{4}{*}{\rotatebox{90}{MSM\cite{kwon2020motionsqueeze}}}  & MSN(reproduced) & TSM-ResNet18 &15  & 11.6  & 46.2 & \multirow{2}{*}{+3.3}\\ 
				~ & ARC-MSN & TSM-ResNet18 &18  & 14.5  & \textbf{49.5} & ~\\  
				\cmidrule{2-7}
				~ & MSN (reproduced) & TSM-ResNet50 &34  & 24.6  & 51.5 & \multirow{2}{*}{+1.5}\\ 
				~ & ARC-MSN & TSM-ResNet50 & 38  & 27.3 & \textbf{53.0} & \\
				\toprule
			\end{tabular} 
		}
	\end{center}
\end{table}

\begin{table}[htbp]
	\begin{center}
		\caption{Performance comparison on Diving48 V2.}
		\label{tab:divingv2}
		\scalebox{0.65}{
			\begin{tabular}{cccccc}   
				\toprule    
				Model & Backbone & GFLOPs$\times{}$clips & Frames & Top-1(\%)\\  
				\toprule 
				
				%
				SlowFast16x8(from \cite{bertasius2021space}) & ResNet101 &213$\times{}$30  & 16$\times{}$30  & 77.6\\ 
				TimeSformer-L\cite{bertasius2021space} & ViT-base &238$\times{}$30  & 16$\times{}$30  & 81.0\\  
				\midrule
				ARC-TSM & ResNet50 &\textbf{74.4$\times{}$2}  & 16$\times{}$2 & \textbf{89.8}\\
				\toprule
			\end{tabular} 
		}
	\end{center}
\end{table}

\section{Overhead Analysis on the ARC Framework}
We quantify several overhead measurements as the functions of the number of recursive steps $n$ in the ARC framework to show its complexity at various aspects. Specifically, a normal convolutional layer is chosen to instantiate the ARC framework. With a slight abuse of notation, the size of feature maps of both the input and the output is set to $C\times{}H\times{}W\times{}T$ and the  kernel size is set to $K\times{}K$. For simplicity, we only count of multipy-add operations here.
\subsection{FLOPs.} Overall, the increase of FLOPs brought by the ARC is relatively small compared to the baseline models. And it is almost a constant with respect to $n$. The number of FLOPs consists of three parts. The first part is exactly from the FLOPs of the convolutional layer itself, i.e., $K^2C^2HWT$. The second part comes from the embedding projection and its FLOPs is $C^2HWT$. The third part is introduced by the backward attention and adaptive fusion, i.e., $\frac{C^2(HW+T)}{n}$. Note that the third part will be executed for $n-1$ times recursively. When $n>1$, the FLOPs of an ARC-equipped convolutional layer is
\begin{equation}
\begin{split}
FLOPs = &K^2C^2HWT+C^2HWT+\\
&\frac{C^2(HWT+HW+T)}{n}(n-1)
\end{split}
\end{equation}

\subsection{Number of Parameters.} Similar to the case of FLOPs, the increase of parameters brought by ARC is relatively small. It is also almost a constant with respect to $n$ and consists of three parts. When $n>1$, for an ARC-equipped convolutional layer,  the number of its parameters is
\begin{equation}
\begin{split}
\#Parameters = &K^2C^2+C^2+\frac{C^2}{n}(n-1)
\end{split}
\end{equation}

\subsection{Peak Memory.} The peak memory is the main obstacle to increasing $n$, as the evolving states at every iterations should be preserved in the memory and thus the peak memory grows linearly with respect to $n$ in training phase. It consists of the original feature map and the embedded feature map at full resolution, the evolving states at full resolution, the spatial attention map  and the temporal attention map:
\begin{equation}
\label{mem_func}
\begin{split}
\#Memory = & 2C^2HWT + (n-1)C^2HWT +\\
& CHW + CT
\end{split}
\end{equation}

\clearpage


